\documentclass[letterpaper]{article} 
\usepackage{aaai2027} 
\usepackage[hyphens]{url}  
\usepackage{graphicx} 
\usepackage{natbib}  
\usepackage{caption} 
\usepackage{algorithm}
\usepackage{algorithmic}

\usepackage{newfloat}
\usepackage{listings}
\DeclareCaptionStyle{ruled}{labelfont=normalfont,labelsep=colon,strut=off} 
\floatstyle{ruled}
\newfloat{listing}{tb}{lst}{}
\floatname{listing}{Listing}

\usepackage{booktabs}

\usepackage{xspace} 
\usepackage{subcaption} 
\usepackage{amsmath} 
\usepackage{amsthm} 
\usepackage{amssymb} 
\usepackage{mathrsfs} 
\usepackage{textcomp} 
\usepackage{diagbox} 
\usepackage{listings} 
\usepackage{tcolorbox} 
\usepackage{multirow} 
\usepackage{tabularx}    
\usepackage{array}       
\usepackage{makecell}    
\usepackage{enumitem} 

\usepackage{colortbl} 

\usepackage{longtable} 
\usepackage{pifont} 

\newcommand{\myparagraph}[1]{\vspace{1mm} \noindent \textbf{#1.}}

\newcommand{\docevo}{DocTrace\xspace}

\newcounter{researchquestion}

\title{Trace Only What You Need: Structure-Aware On-Demand Hypergraph Memory for Long-Document Question Answering}

\author{
    Xiangjun Zai\textsuperscript{\rm 1}, 
    Xingyu Tan\textsuperscript{\rm 1,\rm 2}, 
    Chen Chen\textsuperscript{\rm 3}, 
    Xiaoyang Wang\textsuperscript{\rm 1}, 
    Wenjie Zhang\textsuperscript{\rm 1}
}
\affiliations{
    \textsuperscript{\rm 1}University of New South Wales,\\
    \textsuperscript{\rm 2}CSIRO,\\
    \textsuperscript{\rm 3}University of Wollongong
    \\
    \{xiangjun.zai, xingyu.tan, xiaoyang.wang1, wenjie.zhang\}@unsw.edu.au, chenc@uow.edu.au

}

\begin{document}

\maketitle

\begin{abstract}
Long-document question answering (QA) requires large language models (LLMs) to reason over evidence scattered across lengthy documents, where answers often depend on event order, section-level context, and cross-part evidence connections. 
Although retrieval-augmented generation (RAG) reduces the input context by retrieving relevant evidence, existing structured RAG methods still face three limitations: costly query-agnostic knowledge organization, insufficient use of original document structure, and no reuse of historical reasoning experience. 
To address these limitations, we propose \textbf{\docevo}, a multi-agent RAG framework for long-document QA that supports query-triggered knowledge organization, document-structure-aware and experience-guided reasoning. 
\docevo preserves document hierarchy with a lightweight document structural tree index, constructs agent-shared hypergraph-structured working memory on demand during reasoning, and stores successful reasoning plans in graph-structured experience memory for future reuse, 
enabling adaptive exploration across related long-document questions. 
Experiments\footnote{Our project page is available at \url{https://zaixjun.github.io/DocTrace/}.} on four long-document QA datasets show that \docevo outperforms the strongest baseline, ComoRAG, with average relative gains of 16.91\% in F1 and 15.50\% in EM on open-form QA benchmarks, and by up to 20.67\% and 23.78\%, respectively, on NarrativeQA.

\end{abstract}

\definecolor{qblue}{RGB}{30,90,180}
\definecolor{trgreen}{RGB}{0,140,60}
\definecolor{dmpurple}{RGB}{120,70,170}
\definecolor{wmorange}{RGB}{210,110,20}
\definecolor{agent}{RGB}{0,130,130}

\newcommand{\Badge}[2][black]{%
  \begingroup
  \setlength{\fboxsep}{1.5pt}%
  \fcolorbox{#1}{white}{\textcolor{#1}{\scriptsize\sffamily\bfseries #2}}%
  \endgroup
}

\newcommand{\QTool}[1]{%
  \ensuremath{\mathop{\text{\Badge[qblue]{#1}}}\nolimits}%
}

\newcommand{\WMTool}[1]{%
  \ensuremath{\mathop{\text{\Badge[wmorange]{#1}}}\nolimits}%
}

\newcommand{\TreeTool}[1]{%
  \ensuremath{\mathop{\text{\Badge[trgreen]{#1}}}\nolimits}%
}

\newcommand{\AgentTool}[1]{%
  \ensuremath{\mathop{\text{\Badge[agent]{#1}}}\nolimits}%
}

\newcommand{\DMTool}[1]{%
  \ensuremath{\mathop{\text{\Badge[dmpurple]{#1}}}\nolimits}%
}

\newcommand{\PlainTool}[1]{%
  \ensuremath{\mathop{\text{\sffamily\bfseries #1}}\nolimits}%
}

\section{Introduction}\label{sec:intro}

Long-document question answering (QA) aims to answer questions over lengthy documents, such as books, reports, and narratives~\cite{DBLP:conf/acl/bai2025longbenchv2}. 
A straightforward way to apply large language models (LLMs) to this setting is to prompt the entire document as input context to generate the answer directly~\citep{DBLP:conf/icml/2025lara}. 
Although this full-context strategy can retrieve explicitly stated facts, 
prior works show that LLMs still struggle to reason with information distributed across long contexts, resulting in degraded QA performance~\cite{DBLP:conf/icml/modarressi2025nolima,DBLP:conf/acl/jin2025longrefiner}.

Retrieval-augmented generation (RAG) provides an alternative solution by separating evidence retrieval from answer generation. 
Instead of prompting the LLM with the full document, RAG retrieves a smaller set of relevant information and uses it as grounded context for generation~\cite{DBLP:conf/icml/2025lara}. 
Traditional RAG, however, organizes documents into flat text chunks and retrieves evidence mainly by semantic similarity.
This design can miss cross-chunk dependencies and weaken the structural relations among distant pieces of evidence~\cite{li2026cam}. 
Such issues are particularly critical in long-document QA, where relevant evidence is often dispersed throughout the document, and accurate answering requires preserving event order or section-level context.

\begin{figure}[t]
    \centering
    \includegraphics[width=1\linewidth]{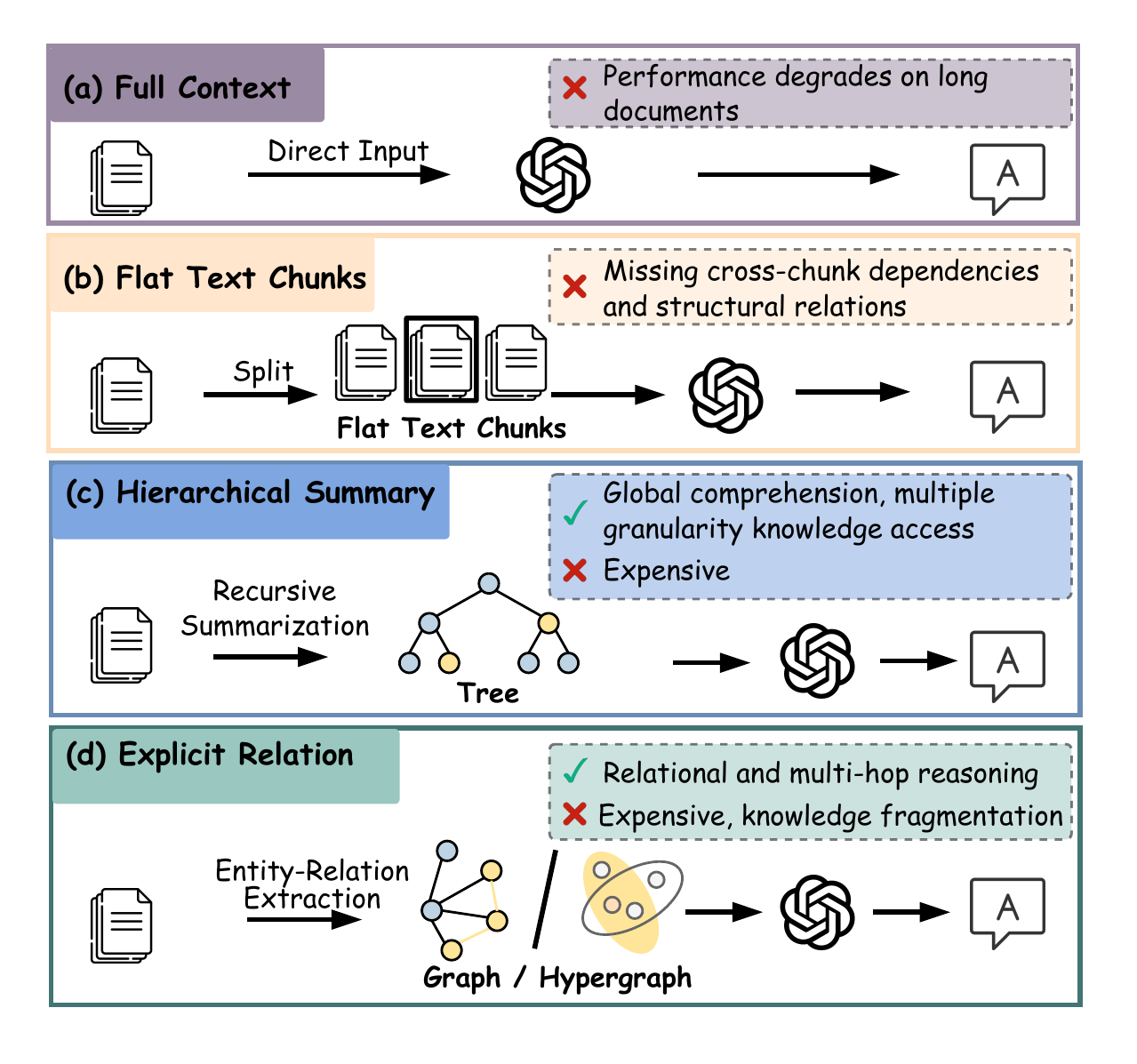}
    \caption{Comparison of four document knowledge organization paradigms for long-document QA.}
   \label{fig:example}
\end{figure}

To address these issues, recent works organize document knowledge in a more structured form, 
as illustrated in Figure~\ref{fig:example}.
One line of work builds hierarchical summary structures. 
For example, RAPTOR~\cite{DBLP:conf/iclr/RAPTORSarthiATKGM24} constructs 
a summary tree that captures different levels of abstraction over document chunks. This approach improves global comprehension and supports knowledge access at multiple granularities.
Another line of work represents document content through explicit relations. 
Methods such as GraphRAG~\cite{edge2024local}, HippoRAG2~\cite{jimenez2025hipporag2}, and HyperGraphRAG~\cite{luo2025hypergraphrag} extract entities and relations from document chunks to construct knowledge graphs or hypergraphs. 
This approach makes relations explicit and facilitates relational and multi-hop reasoning. 
However, because they operate on fine-grained extracted units, globally coherent information may be fragmented across many graph elements~\cite{DBLP:conf/emnlp/tao2025saki}.

\myparagraph{Limitations of existing methods}
Recent hybrid structured methods aim to combine the strengths of hierarchical and relational organization. 
For example, 
ComoRAG~\cite{DBLP:conf/aaai/comorag} organizes document knowledge into multiple complementary structured layers and demonstrates strong performance on long narrative reasoning tasks. 
However, existing structured RAG methods have three unresolved limitations.

\noindent
\underline{L1: Query-agnostic knowledge organization.}
Many existing structured RAG methods first index the entire document into a summary tree or knowledge graph before answering any questions. 
Because the index is constructed independently of the query and the current reasoning state, a large portion of the indexed knowledge may have limited downstream utility. 
Moreover, LLM-based recursive summarization and relation extraction incur high upfront latency and token costs on long documents~\cite{xiang2026whengraph}, making full structured indexing inefficient when only a small portion is question-relevant.

\noindent
\underline{L2: Lack of document structure utilization.}
Within a long document, passages contain not only local factual content, but also contextual meaning determined by their structural position~\cite{DBLP:conf/acl/jin2025longrefiner}. 
For example, the same question may have different answers across sections as events progress, and adjacent paragraphs jointly preserve narrative chronology~\cite{DBLP:conf/aaai/comorag}.
However, current methods split documents into flat chunks before cluster-based summarization or relation extraction, weakly preserving section boundaries, paragraph order, and local neighborhood relations.

\noindent

\noindent
\underline{L3: No reuse of historical reasoning experience.}
Complex questions over long documents often require multi-step retrieval and reasoning, where later retrieval decisions depend on earlier intermediate results~\cite{DBLP:conf/acl/trivedi2023ircot,DBLP:conf/aaai/comorag}.
However, most existing systems treat each question as an isolated instance.
Once a question is answered, the resulting reasoning trajectories are usually discarded. 
Consequently, when structurally similar questions appear later, they must repeat similar exploration from scratch, instead of reusing successful reasoning patterns from prior questions.

\myparagraph{Contribution}
To address these limitations, we introduce \textbf{\docevo}, a multi-agent RAG framework for long-document QA, with an orchestrator that plans and controls the reasoning process, and investigator agents that retrieve evidence and answer subquestions.
\docevo enables document-structure-aware retrieval, integrates document content into an agent-shared working memory on demand, traces evidence within this working memory, and generates answers guided by both the content of the evidence and its position in the document structure.
Advantages of \docevo can be summarized as:

\noindent\raisebox{0.2ex}{\scriptsize$\bullet$}\enspace
\textbf{Query-triggered knowledge organization.}
To avoid the cost of structured knowledge organization, \docevo incrementally constructs a hypergraph-structured working memory $\mathcal{M}$ from document regions selected to meet subquestions' evidence needs.
$\mathcal{M}$ contains two complementary views: an abstraction view $\mathcal{M}_{A}$ for reusable hierarchical summaries bounded by document structure, and a relation view $\mathcal{M}_{R}$ for n-ary relations.
$\mathcal{M}$ is shared across agents and persists across subquestions, enabling later subquestions to reuse organized knowledge without repeated construction.

\noindent\raisebox{0.2ex}{\scriptsize$\bullet$}\enspace
\textbf{Document-structure-aware retrieval and generation.} 
To better utilize document structure, \docevo builds a lightweight {document structural tree index} $\mathcal{T}$ that preserves the document's inherent hierarchy.
Retrieval over $\mathcal{T}$ leverages structural cues in addition to semantic and lexical similarity, and the tree further guides working memory construction by bounding the scope of summarization and relation extraction.
When generating answers, the positions of retrieved evidence within $\mathcal{T}$ help restore the document's narrative flow.

\noindent\raisebox{0.2ex}{\scriptsize$\bullet$}\enspace
\textbf{Reusable reasoning experience.} 
To allow learning from past reasoning experience, reasoning plans from the orchestrator are explicitly modeled as directed acyclic graphs over subquestions.
High-quality plans are stored in a {graph-structured experience memory} $\mathcal{DM}$ for later reuse on questions with similar reasoning motifs.
Graph traversal in $\mathcal{DM}$ not only supports plan retrieval but also enables composing new plan examples beyond those explicitly stored.

\noindent\raisebox{0.2ex}{\scriptsize$\bullet$}\enspace
\textbf{Effectiveness and adaptability.}
a) Across four long-document QA datasets, \docevo achieves the best performance on all datasets, outperforming ComoRAG by average relative gains of 16.91\% in F1 and 15.50\% in EM on open-form QA, with gains up to 20.67\% and 23.78\%, respectively, on NarrativeQA.
b) \docevo is plug-and-play across diverse document structures and LLM backbones.
By organizing richer document knowledge on demand instead of constructing full summary or relation indexes, it is particularly suitable for newly introduced or sparsely queried documents, where expensive offline preprocessing is difficult to amortize.
c) \docevo further improves through experience reuse.
As reusable reasoning plans accumulate, the benefit of experience memory shows growth trends.

\section{Related Work}
\label{sec:related_work}

\begin{figure*}[t]
    \centering
    \includegraphics[width=1.0\linewidth]{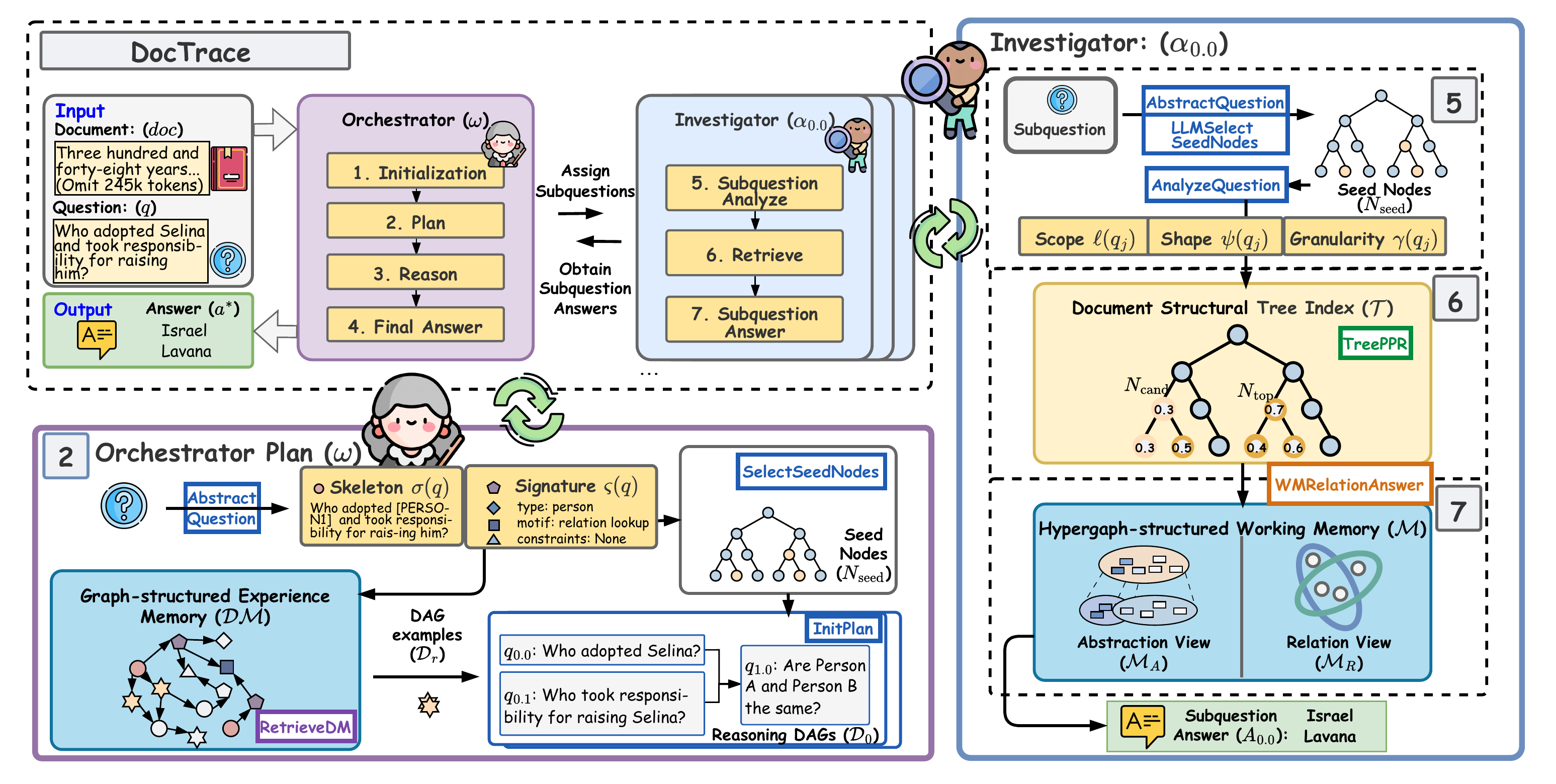}
    \caption{Overview of the \docevo framework. \docevo adopts an \textbf{Orchestrator-Investigator Architecture} that performs reasoning, retrieval, and answer generation guided by a \textbf{Document Structural Tree Index~$\mathcal{T}$}. It maintains a query-triggered \textbf{Hypergraph-structured Working Memory $\mathcal{M}$} to integrate document content and a cross-query \textbf{Graph-structured Experience Memory}~$\mathcal{DM}$ for the Orchestrator to reuse successful reasoning plans.}
    \label{fig:pipeline}
\end{figure*}

\myparagraph{Structured RAG}
Recent RAG methods move beyond flat-chunk retrieval by organizing knowledge into structured indexes, 
including knowledge graphs~\cite{edge2024local,guo-etal-2025-lightrag,DBLP:conf/nips/jimenez2024hipporag,jimenez2025hipporag2,DBLP:conf/kdd/ketrag2025}, hypergraphs~\cite{luo2025hypergraphrag,feng2026hyperrag,zhou2025improvingmultistepraghypergraphbased,zai2025proh}, and hierarchical summary trees~\cite{DBLP:conf/iclr/RAPTORSarthiATKGM24,li2026cam}. 
Hybrid systems such as ComoRAG~\cite{DBLP:conf/aaai/comorag} integrate multiple structured layers into the index.
However, most existing methods build the full structured index before receiving a query, incurring substantial upfront cost and producing summaries, entities, or relations with little query utility~\cite{xiang2026whengraph,zhuang2025linearrag}.

\myparagraph{Document structure-aware retrieval}
Long documents naturally contain hierarchical and sequential structures, such as section boundaries, paragraph order, and local neighborhoods. 
These structural signals provide contextual meaning beyond isolated chunks and can help locate and interpret evidence~\cite{DBLP:conf/acl/jin2025longrefiner}. 
Recent methods exploit such signals through discourse-aware representations, tree-based chunking, sliding-window summarization, or multi-granularity retrieval~\cite{chen2025beyond,DBLP:conf/acl/tao2025treerag,wang2025bookrag,DBLP:conf/aaai/htsir2025,DBLP:conf/acl/jin2025longrefiner}. 

\myparagraph{Planning and memory in retrieval-augmented reasoning}
Complex questions often require multi-step retrieval and reasoning, where later decisions depend on earlier intermediate results. 
General reasoning frameworks, such as ReAct~\cite{DBLP:conf/iclr/yao2023react} and ToT~\cite{DBLP:conf/nips/yao2023ToT}, support dynamic planning through interleaved reasoning and search, while retrieval-oriented methods such as IRCoT~\cite{DBLP:conf/acl/trivedi2023ircot}, Plan*RAG~\cite{verma2024plan}, LogicRAG~\cite{chen2026logicrag} and DeepRAG~\cite{guan2026deeprag} integrate planning with evidence acquisition. 
Recent agent memory methods further study persistent memory for LLM agents~\cite{DBLP:conf/www/qian2025memorag,xu2025amem,zhou2026mem,zhang2025memevolvemetaevolutionagentmemory,rasmussen2025zep,DBLP:conf/acl/cer2025}. 

\section{Method}\label{sec:method}

\newcommand{\circnum}[1]{%
  {\large{\ding{\numexpr201+#1\relax}}}%
}

Given a long document $doc$ and a question $q$, long-document QA aims to generate an answer $a=f(q,doc)$ conditioned on the content of $doc$.
\docevo solves this task with an orchestrator $\omega$ that plans over interdependent subquestions and investigators $\alpha_j$ that retrieve evidence and answer individual subquestions.
As shown in Figure~\ref{fig:pipeline}, the agents operate over three complementary components:
\circnum{1} a lightweight \textbf{Document Structural Tree Index} $\mathcal{T}$ that preserves the document hierarchy;
\circnum{2} an agent-shared, query-triggered \textbf{Hypergraph-structured Working Memory} $\mathcal{M}=(\mathcal{M}_{A},\mathcal{M}_{R})$ that organizes selected evidence through abstraction and relation views; and
\circnum{3} a cross-query \textbf{Graph-structured Experience Memory} $\mathcal{DM}$ that stores reusable reasoning plans represented as directed acyclic graphs (DAGs) of subquestions and dependency relations.
For each question $q$, the orchestrator $\omega$ retrieves relevant DAGs from $\mathcal{DM}$, initializes and refines reasoning DAGs, dispatches subquestions to investigators, synthesizes and selects the final answer, and stores high-quality reasoning DAGs in $\mathcal{DM}$ for later reuse.
We next detail the agent workflow and three components.

\subsection{Orchestrator-Investigator Architecture}\label{sec:method:arch}

Solving complex long-document questions often requires global coordination across interdependent subquestions and targeted evidence retrieval for each.
We therefore adopt an orchestrator-investigator architecture: 
the orchestrator $\omega$ performs question-level planning and coordinates the overall reasoning process, while each investigator $\alpha_j$ executes a subquestion $q_j$ through evidence retrieval and answer generation.
The Document Structural Tree Index $\mathcal{T}$ is built once per document (Section~\ref{sec:method:tree}), whereas for each question $q$, $\omega$ initializes a working memory $\mathcal{M}$ shared by all agents throughout the reasoning process.

\myparagraph{Orchestrator planning}\label{sec:method:arch:omega}
The orchestrator $\omega$ first invokes \QTool{AbstractQuestion}, which uses an LLM to derive three representations of $q$: topic entities $E_q$, a question skeleton $\sigma(q)$, and a question signature $\varsigma(q)$.
These representations serve distinct roles in downstream planning: 
$E_q$ links the question to relevant entities in the document, while $\sigma(q)$ and $\varsigma(q)$ abstract away from document-specific entity identities to support retrieval and reuse of structurally similar reasoning trajectories from Experience Memory~$\mathcal{DM}$.
$\omega$ then invokes \QTool{SelectSeedNodes} to construct the initial document-grounded context for planning by selecting a small set of tree nodes $N_{\mathrm{seed}} \subseteq \mathcal{N}$ likely to contain evidence relevant to $q$.
To perform this selection, \QTool{SelectSeedNodes} scores each tree node by combining three relevance signals:

\begin{equation}\label{eq:seed_score}
s(n, q)
=
\lambda_{\text{sem}} s_{\text{sem}}(n)
+
\lambda_{\text{lx}} s_{\text{lx}}(n)
+
\lambda_{\text{e}} s_{\text{e}}(n),
\end{equation}
where $s_{\text{sem}}(n)$ is semantic similarity score, $s_{\text{lx}}(n)$ is lexical relevance score and $s_{\text{e}}(n)$ is topic entity score.
Using $\sigma(q)$ and $\varsigma(q)$, $\omega$ invokes \DMTool{RetrieveDM} to retrieve related reasoning DAG examples $\mathcal{D}_{r}$ from $\mathcal{DM}$ (Section~\ref{sec:method:dm}).
$\omega$ then invokes \QTool{InitPlan}, which supplies $\mathcal{D}_{r}$, $E_q$, and $N_{\mathrm{seed}}$ as context for an LLM to generate candidate plans comprising subquestions $\mathcal{Q}=\{q_0,\ldots,q_{m-1}\}$ and dependency edges $\mathcal{L}\subseteq\{(i,j)\mid 0\leq i,j<m\}$.
An edge $(i,j)$ indicates that $q_j$ depends on the answer to $q_i$.
Each candidate plan is verified to be a valid DAG using topological sorting, and is accepted as an initial DAG only if its subquestions are relevant, faithful, useful, and collectively cover the core intent of $q$.

\myparagraph{Orchestrator reasoning}
$\omega$ explores active DAG states with a depth-first frontier $\mathcal{F}$.
Initially, $\mathcal{F}=\mathcal{D}_0$, the set of accepted initial plans.
At each iteration, $\omega$ pops one DAG, identifies its earliest unfinished execution level $\mathcal{Q}_i$, and dispatches each subquestion $q_j\in\mathcal{Q}_i$ to an investigator $\alpha_j$.
After inserting the returned level-$i$ answers and evidence into the DAG state, $\omega$ invokes \QTool{LLMGenerateNewDAG}, which prompts an LLM to revise only the unfinished portion of the DAG.
The resulting refined DAGs are then pushed back onto $\mathcal{F}$.
Reasoning terminates as soon as $K$ completed DAGs have been obtained.
These completed DAGs provide the intermediate answers and evidence used for final answer synthesis and selection.

\myparagraph{Investigator retrieval and answering}
For a subquestion $q_j$, investigator $\alpha_j$ first applies \QTool{AbstractQuestion} to derive its question abstraction.
It then invokes \QTool{LLMSelectSeedNodes}, which augments the combined score in Equation~\ref{eq:seed_score} with LLM-based reranking to produce the final seed set $N_{\mathrm{seed}}$.
Next, \QTool{AnalyzeQuestion} predicts three retrieval-geometry variables from the hierarchical structural outline of $\mathcal{T}$ and the positions of $N_{\text{seed}}$:
the scope level $\ell(q_j)$,
retrieval shape $\psi(q_j)$,
and evidence granularity $\gamma(q_j)$.
Based on $\ell(q_j)$, $\psi(q_j)$, and $N_{\mathrm{seed}}$, \TreeTool{TreePPR} then returns two ranked node sets $N_{\mathrm{top}}$ and $N_{\mathrm{cand}}$.
After that, $\alpha_j$ chooses the answering route based on the predicted evidence granularity $\gamma(q_j)$.
For $\gamma(q_j)=\mathsf{relation}$, \WMTool{WMRelationAnswer} constructs or reuses the Relation View $\mathcal{M}_{R}$ over $N_{\mathrm{top}}$ and answers using relational paths.
Otherwise, an adaptive text-reading route progressively widens a tree node set $N_k$ (initialized as $N_{\mathrm{top}}$) by adding nodes from $N_{\mathrm{cand}}$.
If $N_k$ does not fit within the direct-read budget, $\alpha_j$ calls \WMTool{WMAbstractAnswer} to construct or reuse the Abstraction View $\mathcal{M}_{A}$; otherwise, it calls \WMTool{DirectAnswer}, assembling the selected nodes in their original document order to form the answer context.
If all widening attempts fail, $\alpha_j$ falls back to \WMTool{DirectAnswer} over $N_{\mathrm{top}} \cup N_{\mathrm{cand}}$.

\subsection{Document Structural Tree Index}\label{sec:method:tree}

Passages in long documents contain both local facts and contextual meanings determined by their structural positions.
We therefore represent each document as a structural tree rather than a flat chunk list, allowing retrieval to expand, bound, and order evidence according to document organization.
Formally, the Document Structural Tree is $\mathcal{T}=(\mathcal{N},\mathcal{E}_n,\mathcal{L}_t,\mathcal{L}_n)$, where $\mathcal{N}$ contains document nodes, $\mathcal{E}_n$ contains their named entities, $\mathcal{L}_t$ contains tree edges, and $\mathcal{L}_n$ links nodes to entities.
Each node $n\in\mathcal{N}$ belongs to the $\mathsf{document}$, $\mathsf{chapter}$, $\mathsf{section}$, or $\mathsf{paragraph}$ level. 
$n$ stores its text $x_n$ and an outline-style preview of its children.
To support fast retrieval, we maintain vector and BM25 indexes over node content.

\myparagraph{Document tree construction}
$\mathcal{T}$ is constructed using a two-tier strategy.
\TreeTool{BuildDocTree} first extracts structural cues and representative content previews.
It then uses an LLM to infer the document heading pattern, parsing the document into nodes $\mathcal{N}$ and constructing the corresponding tree edges $\mathcal{L}_t$.
If $(\mathcal{N}, \mathcal{L}_t)$ fails structural or quality checks, it falls back to an LLM-generated parsing function for tree construction.
Finally, named entity recognition (NER) is applied to each node using a lightweight spaCy model~\cite{Honnibal_spaCy_Industrial-strength_Natural_2020}, producing entity set $\mathcal{E}_n$ and node-entity links $\mathcal{L}_n$.

\myparagraph{TreePPR}
To propagate seed relevance through document structure, \TreeTool{TreePPR} applies Personalized PageRank (PPR) to a temporary directed graph containing parent-child tree edges within scope $\ell$ and bidirectional sibling edges.
Edge weights are assigned according to retrieval shape $\psi$ to support different evidence geometries.
It computes the stationary personalized relevance vector $\rho$ as
\begin{equation}\label{eq:tree_ppr}
\rho
=
\alpha_{\psi} P_{\ell,\psi}^{\top}\rho
+
(1-\alpha_{\psi}) v,
\end{equation}
where $P_{\ell,\psi}$ is the row-stochastic, retrieval-geometry-conditioned transition matrix, $v$ is the normalized seed distribution, and $\alpha_{\psi}$ is the damping coefficient.
Finally, ranking tree nodes $\mathcal{N}$ by $\rho$ produces the node sets $N_{\mathrm{top}}$ and $N_{\mathrm{cand}}$.

\subsection{Hypergraph-Structured Working Memory}
\label{sec:method:wm}
Questions may require evidence at different levels of abstraction: some require local factual details, some require higher-level semantic abstraction, and others require tracing relational connections among entities across passages.
However, constructing a summary hierarchy or a knowledge graph of the entire document is computationally expensive. 
We therefore allow agents to construct a dual-view Working Memory $\mathcal{M}=(\mathcal{M}_{A},\mathcal{M}_{R})$ dynamically in a query-triggered manner, and share it across subquestions.
In abstraction view $\mathcal{M}_{A}$, hyperedges group child summaries or tree nodes, while in relation view $\mathcal{M}_{R}$, hyperedges encode n-ary relations among entities.
Both views retain links to source tree nodes, allowing evidence to be traced to the original document.

\myparagraph{Abstraction view}
The abstraction view $\mathcal{M}_{A}$ supports answering subquestions when evidence is too broad for direct reading. It constructs reusable hierarchical summaries over selected regions of $\mathcal{T}$.
For node set $N_k$ and target granularity $\gamma$, $\mathcal{M}_{A}$ builds bottom-up summaries until reaching $\gamma$.
It orders nodes by their positions in $\mathcal{T}$, groups siblings under their parents, and progressively merges these groups while preserving document position.
Each summarization group contains at most $W_{\mathrm{sum}}$ nodes.
When an overlapping and semantically coherent summary already exists, $\mathcal{M}_{A}$ incrementally updates the existing summary instead of reconstructing it from scratch.
\WMTool{WMAbstractAnswer} answers $q_j$ through a three-tier procedure.
It first retrieves summaries that structurally cover $N_k$ at granularity $\gamma$, then retrieves summaries at the same granularity based on their semantic similarity to $q_j$.
If they cannot answer $q_j$, $\mathcal{M}_{A}$ constructs fresh summaries over $N_k$ and uses them to generate an answer.
During abstraction-based answering, the LLM first answers from the summaries and may request a drill-down to specific original tree nodes when the abstraction is insufficient.

\myparagraph{Relation view}
$\mathcal{M}_{R}$ supports explicit multi-hop reasoning over n-ary relations. 
Rather than building a document-wide knowledge hypergraph in advance, $\mathcal{M}_{R}$ incrementally extracts n-ary relations $\mathcal{R}_j$ and participating entities $\mathcal{E}_j$ only from selected tree nodes $N_{\text{top}}$ for the current subquestion $q_j$.
$\mathcal{R}_j$ and $\mathcal{E}_j$ are retained in $\mathcal{M}_{R}$ with links to their source tree nodes for reuse by subsequent subquestions.
N-ary relation extraction and reasoning-path retrieval follow prior work~\cite{luo2025hypergraphrag,zai2025proh}.
To answer $q_j$, \WMTool{WMRelationAnswer} forms a scoped hypergraph $\mathcal{H}=(\mathcal{E}_H,\mathcal{R}_H)$ from the relations and entities associated with $N_{\text{top}}$. 
$\mathcal{H}$ is further expanded with additional tree nodes that are strongly associated with the topic entities $E_{qj}$. 
\WMTool{WMRelationAnswer} then retrieves supporting reasoning paths from $\mathcal{H}$ and generates $A_j$ from them.

\subsection{Graph-Structured Experience Memory}\label{sec:method:dm}

\begin{table*}[t]
\centering
\small
\begin{tabular}{ll|cccccc}
\toprule
\multirow{2}{*}{\textbf{Category}} &
\multirow{2}{*}{\textbf{Method}} &
\multicolumn{2}{c}{\textbf{NarrativeQA}} &
\multicolumn{2}{c}{\textbf{EN.QA}} &
\textbf{EN.MC} &
\textbf{DetectiveQA} \\
\cmidrule(lr){3-4}
\cmidrule(lr){5-6}
\cmidrule(lr){7-7}
\cmidrule(lr){8-8}
& & F1 & EM & F1 & EM & ACC & ACC \\
\midrule
\multirow{1}{*}{\shortstack[c]{\textbf{LLM Prompting}}}
& Full context + IO
& 16.28 & 10.60
& 11.97 & 8.26
& 24.89
& 35.00 \\
\midrule
\multirow{2}{*}{\shortstack[c]{\textbf{Text-based RAG}}}
& Flat RAG
& 26.22 & 14.20
& 21.41 & 14.81
& 52.84
& 50.00 \\
& LogicRAG~\cite{chen2026logicrag}
& 30.27 & 16.60
& 26.67 & 16.24
& 40.61
& 40.00 \\
\midrule
\multirow{2}{*}{\shortstack[c]{\textbf{Summary-based RAG}}}
& RAPTOR~\cite{DBLP:conf/iclr/RAPTORSarthiATKGM24}
& 27.44 & 15.00
& 21.42 & 15.38
& 56.77
& 49.17 \\

& RAPTOR+IRCoT
& 31.94 & 18.40
& 24.02 & 16.52
& 60.26
& 48.33 \\
\midrule
\multirow{2}{*}{\shortstack[c]{\textbf{Graph-based RAG}}}
& HippoRAG2~\cite{jimenez2025hipporag2}
& 26.62 & 14.00
& 19.98 & 13.96
& 58.95
& 49.17 \\

& HippoRAG2+IRCoT
& 32.32 & 17.20
& 29.67 & 21.08
& 65.79
& 51.95 \\

\midrule
\multirow{3}{*}{\shortstack[c]{\textbf{Hybrid Structured RAG}}}
& ComoRAG~\cite{DBLP:conf/aaai/comorag}
& \underline{34.54} & \underline{18.80}
& \underline{34.67} & {24.88}
& 61.43
& 41.39 \\
& ComoRAG~(Originally Reported)
& 31.43 & 18.60
& {34.52} & \underline{25.07}
& \underline{72.93}
& \underline{68.18} \\
& ComoRAG~(Best)
& \underline{34.54} & \underline{18.80}
& \underline{34.67} & \underline{25.07}
& \underline{72.93}
& \underline{68.18} \\
\midrule

\shortstack[l]{\textbf{Proposed}\\ \textbf{Method}}
& \shortstack[l]{\textbf{\docevo}\\ Relative Gain (+\%)}
& \shortstack[c]{\textbf{41.68}\\20.67\%}
& \shortstack[c]{\textbf{23.27}\\23.78\%}
& \shortstack[c]{\textbf{39.22}\\13.12\%}
& \shortstack[c]{\textbf{26.88}\\7.22\%}
& \shortstack[c]{\textbf{76.86}\\5.39\%}
& \shortstack[c]{\textbf{69.72}\\2.26\%} \\
\bottomrule

\end{tabular}
\caption{Results on four long-document QA datasets. The \textbf{best} and \underline{second-best} results are highlighted. \textbf{All results are reproduced using GPT-4o-mini}. ``(Best)'' reports the better result between the reproduced and originally reported scores. Percentage values below the \docevo line report the relative gains of \docevo over ComoRAG~(Best).}
\label{tab:main}
\end{table*}

\docevo solves complex questions by structurally decomposing them into simpler subquestions, which form a reasoning DAG. A successful DAG not only solves the current question, but also provides reusable experience for future questions. This experience records which intermediate information should be retrieved, how they depend on each other, and which reasoning structure leads to a valid answer. Since structurally similar questions may require similar decomposition even when their entities and source documents differ, \docevo stores successful DAGs in a graph-structured memory for reuse across questions.

\myparagraph{Memory schema}
$\mathcal{DM}$ is a heterogeneous graph consisting of six node types: skeleton $\sigma$, reasoning DAG $D$, signature description $\varsigma_{des}$, reasoning motif $\varsigma_m$, answer type $\varsigma_t$, and constraint $\varsigma_c$. 
Edges in $\mathcal{DM}$ link a reasoning DAG $D$ to both the parent skeleton (the original question) and the child skeletons (the subquestions). This parent-plan-child structure supports parent-child DAG composition during retrieval.
$\mathcal{DM}$ also includes edges connecting signature descriptions to skeletons, reasoning motifs, answer types, and constraints. 
These edges provide alternative retrieval routes for relevant DAGs.

\myparagraph{Retrieval}
Given the abstraction $(\sigma(q),\varsigma(q))$ of question $q$, \DMTool{RetrieveDM} identifies candidate DAG examples from three retrieval paths emphasizing different aspects: semantic similarity of the question skeleton $\sigma$, semantic similarity of the reasoning signature description $\varsigma_{des}$, and structural overlap in reasoning motifs, answer types, and constraints. 
The candidate DAGs are then scored, ranked and selected based on these aspects, together with a stored DAG quality score.
For each selected DAG, $\mathcal{DM}$ expands a child subquestion using its own DAG, thereby dynamically composing a new DAG example beyond those explicitly stored in $\mathcal{DM}$.

\myparagraph{Update}
To improve the quality and utility of stored experiences, $\mathcal{DM}$ accepts only high-quality reasoning DAGs.
Given a completed DAG, \DMTool{UpdateDM} first evaluates its quality based on the faithfulness, coverage, and structural validity of its decomposition, together with the coherence between the subquestion answers and the final answer.
For an accepted DAG, $\mathcal{DM}$ stores the reasoning DAG together with its associated parent and child skeletons. Each skeleton is resolved to a canonical node to avoid storing duplicate skeletons.
$\mathcal{DM}$ then inserts corresponding signature components associated with each skeleton.
To bound memory growth, $\mathcal{DM}$ retains only top-scored DAGs under each skeleton and limits the number of skeletons associated with each reasoning motif.

\section{Experiment}
\label{sec:experiment}

\myparagraph{Datasets \& Metrics}
Following prior work~\cite{DBLP:conf/aaai/comorag}, we evaluate \docevo on four commonly used long-document QA datasets: DetectiveQA~\cite{xu2024detectiveqa}, NarrativeQA~\cite{DBLP:journals/tacl/KociskyNarrativeQA2018}, and the EN.MC and EN.QA of $\infty$BENCH~\cite{zhang-etal-2024-bench}. These datasets span diverse document structures and include both open-form and multiple-choice QA tasks, with average document lengths ranging from approximately 100k to 200k tokens for different levels of difficulty in long-document understanding.
We report F1 and Exact Match (EM) for open-form QA to capture partial and exact answer correctness, respectively, and accuracy (ACC) for multiple-choice QA.

\myparagraph{Baselines}
We compare \docevo to
five categories of baselines: (1) {LLM Prompting}, which prompts the LLM directly with the document within a 120k-token context; (2) {Text-based RAG}, including Flat RAG, which retrieves 512-token chunks, and LogicRAG, which guides flat passage retrieval with a query-side logic dependency DAG; (3) {Summary-based RAG}, including RAPTOR and RAPTOR+IRCoT; (4) {Graph-based RAG}, including HippoRAG2, HippoRAG2+IRCoT; and (5) {Hybrid structured RAG}, represented by ComoRAG.
The IRCoT variants provide iterative-retrieval comparisons with \docevo.

\myparagraph{Evaluation protocol}
Baseline implementations and hyperparameters follow the corresponding official codebases. Unless otherwise specified, all methods use GPT-4o-mini as the LLM backbone, BGE-M3~\cite{chen-etal-2024-m3} for embeddings, and identical answer-generation prompts. For multiple-choice QA, answer options are withheld during retrieval and provided only at the final answer-generation stage. Under the online experience-memory protocol, we initialize $\mathcal{DM}$ as empty and process questions sequentially in ascending order of question ID. 

\subsection{Main Results}

We compare \docevo against eight reproduced baselines and report results averaged over three runs. For ComoRAG, we include both reproduced and originally reported results, with ComoRAG~(Best) as a strong reference. Despite this conservative comparison, \docevo achieves the best performance on all datasets, as shown in Table~\ref{tab:main}.
For open-form QA, \docevo achieves 41.68 F1 and 23.27 EM on NarrativeQA, and 39.22 F1 and 26.88 EM on EN.QA. These results represent average relative gains of 16.91\% in F1 and 15.50\% in EM over ComoRAG~(Best), reaching 20.67\% and 23.78\%, respectively, on NarrativeQA.
For multiple-choice QA, \docevo achieves 76.86 ACC on EN.MC and 69.72 ACC on DetectiveQA, still outperforming ComoRAG.
These consistent gains across open-form and multiple-choice QA demonstrate generalization across answer formats. Moreover, \docevo requires no preconstructed document-level graph or summary index.

\subsection{Ablation Studies}

\begin{figure}[tb]
    \centering
    \includegraphics[width=1.0\linewidth]{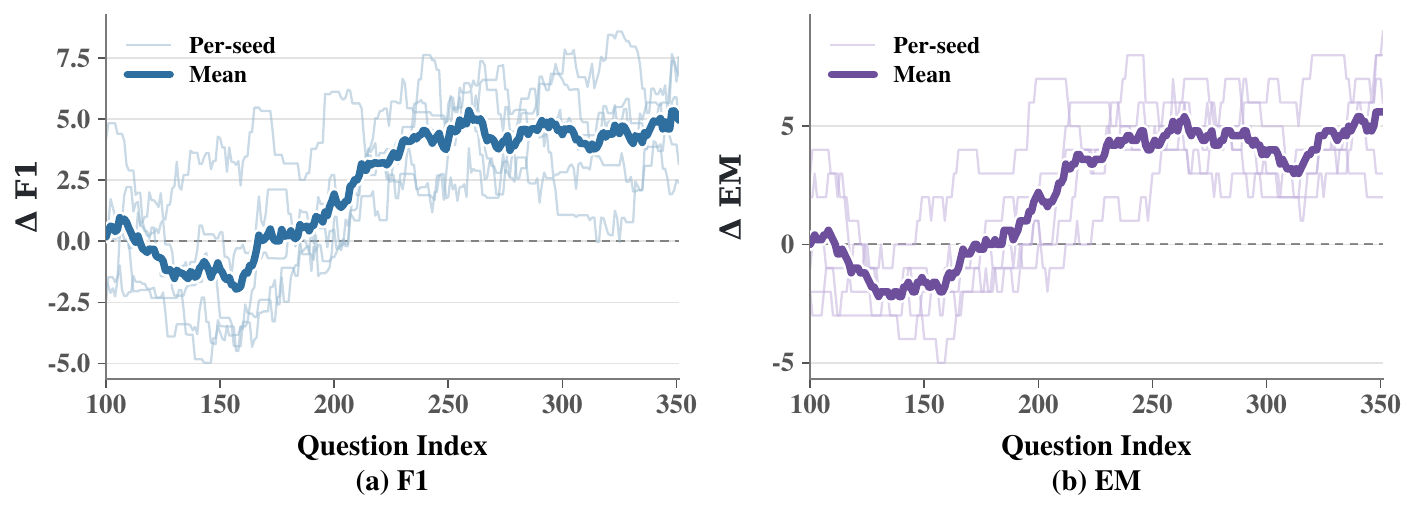}
    
    \caption{Growth of the Experience Memory benefit on EN.QA across five question orders. The curves show trailing 100-question means of the paired performance differences between the Cold-Online and No-$\mathcal{DM}$ settings.}\label{fig:ol_enqa}

\end{figure}

\myparagraph{Ablation on experience memory}
We examine whether the Experience Memory $\mathcal{DM}$ benefits \docevo and whether its benefit grows as reusable reasoning experience accumulates.
Using five random seeds, we generate five question orders on EN.QA.
For each order, we run \docevo in the Cold-Online setting, starting from an empty $\mathcal{DM}$.
To control for variation in question difficulty along each order, we pair every Cold-Online score with the No-$\mathcal{DM}$ score for the same question and analyze their difference.
Figure~\ref{fig:ol_enqa} visualizes the paired differences using trailing 100-question means.
For both F1 and EM, the mean paired advantage begins near zero, dips below zero during the early stage, and then increases as more questions are processed.
To quantify this trend, we fit an ordinary least squares (OLS) slope to the raw paired differences within each question order.
All five question orders exhibit positive slopes for both F1 and EM.
These results demonstrate that $\mathcal{DM}$ improves \docevo and that its benefit increases as reusable reasoning experiences accumulate.

\myparagraph{Ablation on working memory and document structure}
To isolate the contributions of the remaining core components, we ablate the Working Memory $\mathcal{M}$ and Document Structural Tree Index $\mathcal{T}$ individually and jointly on NarrativeQA and DetectiveQA.
As shown in Table~\ref{tab:ablation_study}, removing $\mathcal{M}$ decreases NarrativeQA F1 and EM by 4.64 and 6.27 points, respectively, and DetectiveQA ACC by 4.72 points.
This confirms that query-triggered abstraction and relation views help organize and integrate evidence at different granularities.
Removing $\mathcal{T}$ causes drops of 4.07 F1 and 6.77 EM on NarrativeQA and 8.89 ACC on DetectiveQA, demonstrating the importance of document hierarchy and ordering for locating and interpreting evidence across long documents.
Removing both components produces the largest overall degradation, reducing NarrativeQA F1 by 6.81 points and DetectiveQA ACC by 12.22 points.
These results demonstrate the combined effectiveness of $\mathcal{T}$ and $\mathcal{M}$: $\mathcal{T}$ identifies and orders relevant document regions, while $\mathcal{M}$ organizes their content for multi-granularity reasoning.

\begin{table}[t]
\centering
\small
\begin{tabular}{lccc}
\toprule
\multirow{2}{*}{\textbf{Method}} & \multicolumn{2}{l}{\textbf{NarrativeQA}} & \multicolumn{1}{c}{\textbf{DetectiveQA}} \\
\cmidrule(lr){2-3}\cmidrule(lr){4-4}
 & F1 & EM & ACC \\
\midrule
\textbf{\docevo} & \textbf{41.68} & \textbf{23.27}
& \textbf{69.72} \\
~~w/o $\mathcal{M}$ & 37.04 & 17.00 & 65.00 \\
~~w/o $\mathcal{T}$ & 37.61 & 16.50 & 60.83 \\
~~w/o both & 34.87 & 16.50 & 57.50 \\
\bottomrule
\end{tabular}
\caption{Ablation studies of {\docevo} components.}
\label{tab:ablation_study}
\end{table}

\begin{figure}[t]
    \centering
    \includegraphics[width=1.0\linewidth]{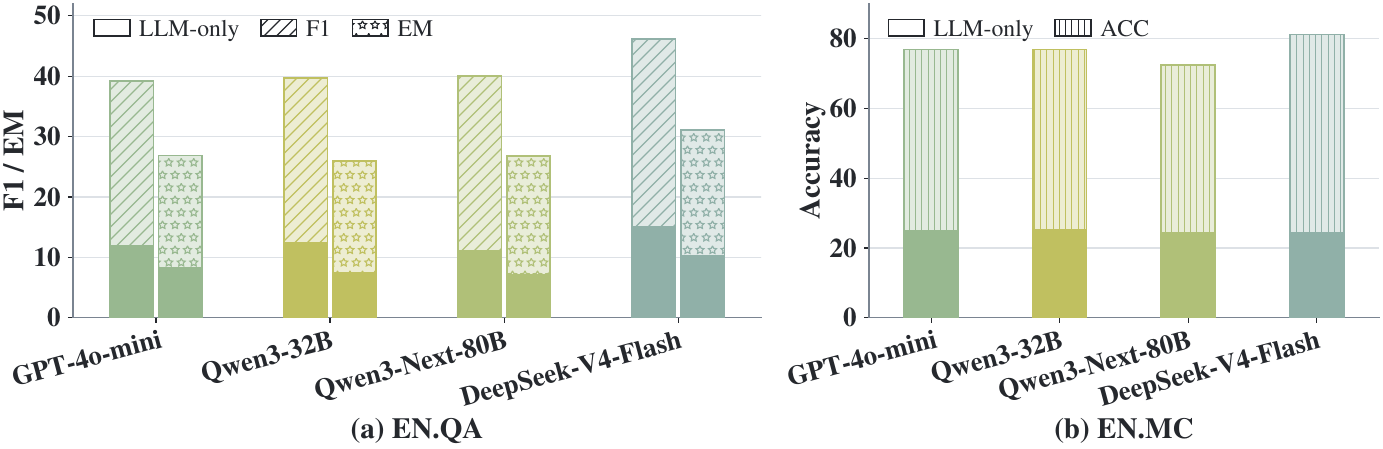}
    \caption{Performance of \docevo and IO prompting across different LLM backbones.}\label{fig:vary_llm}
\end{figure}

\myparagraph{LLM backbone effect}
We evaluate \docevo using four LLM backbones (GPT-4o-mini, Qwen3-32B, Qwen3-Next-80B, and DeepSeek-V4-Flash) on EN.QA and EN.MC.
We also compare \docevo against direct IO prompting with the same LLM backbone, where the entire document is provided to the model under a 120k-token context limit.
As shown in Figure~\ref{fig:vary_llm}, stronger backbones generally further improve the performance of \docevo. 
DeepSeek-V4-Flash achieves the best results on both datasets, reaching an F1 of 46.12 and an EM of 31.05 on EN.QA, and an ACC of 81.22 on EN.MC.
Across all four backbones, \docevo consistently outperforms direct IO prompting, with average absolute gains of 28.64, 19.40, and 52.08 points in F1, EM, and ACC, respectively.

\subsection{Effectiveness Evaluation}

\begin{figure}[t]
    \centering
    \includegraphics[width=\linewidth]{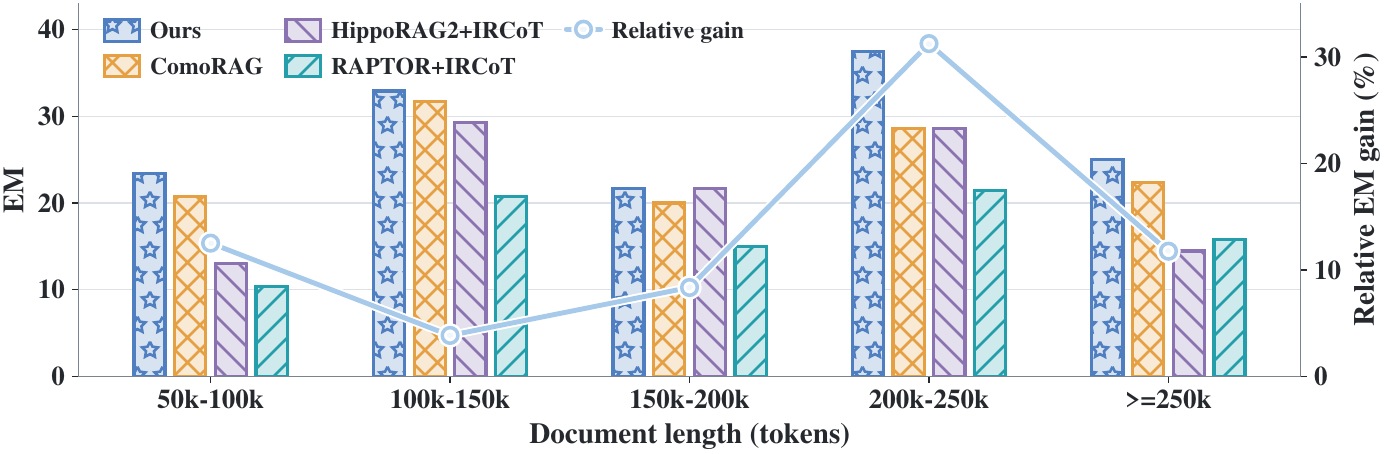}
    \caption{EM on EN.QA across document-length buckets. The line reports the relative EM gain of \docevo over ComoRAG within each bucket.}
    \label{fig:length_performance}
\end{figure}

\myparagraph{Performance by document length}
We evaluate \docevo across document-length ranges on EN.QA and compare it with ComoRAG, HippoRAG2+IRCoT, and RAPTOR+IRCoT using EM.
As shown in Figure~\ref{fig:length_performance}, \docevo ranks first or tied first in all five ranges.
Its largest advantage occurs in the 200k--250k range, where it reaches 37.50\% EM and outperforms the strongest baselines by 8.93 points, while remaining above them beyond 250k tokens.
Although absolute EM varies across buckets because they contain different documents and questions with varying difficulty, \docevo ranks first or tied first in every range.
We further compare \docevo with ComoRAG within each bucket.
\docevo achieves positive relative EM gains across all document-length ranges, with the gain reaching 31.25\% in the 200k--250k range and remaining positive for documents beyond 250k tokens.
These results show that \docevo's advantage extends beyond shorter inputs and is particularly clear above 200k tokens, demonstrating the value of document-structure-aware, query-triggered reasoning for long-document understanding.

\begin{figure}[t]
    \centering
    \includegraphics[width=\linewidth]{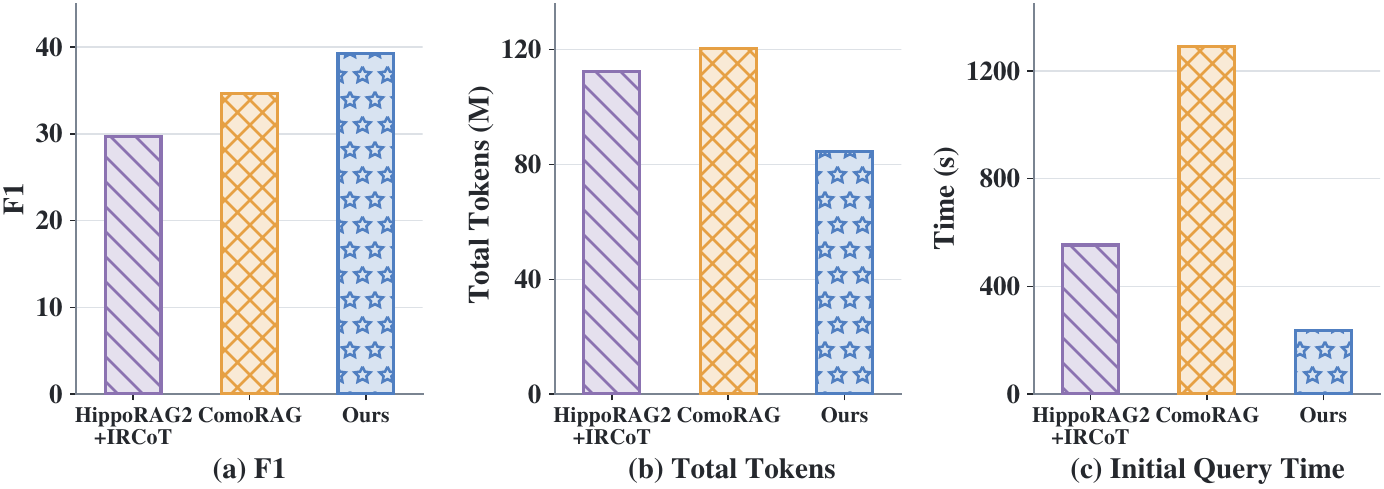}
    \caption{Efficiency comparison of \docevo and the baselines on EN.QA.}
    \label{fig:performance_efficiency}
\end{figure}

\subsection{Efficiency Analysis}

\myparagraph{Computational efficiency}
We compare the token usage and latency of \docevo with HippoRAG2+IRCoT and ComoRAG on EN.QA.
As shown in Figure~\ref{fig:performance_efficiency}, \docevo achieves the highest F1 score of 39.22 with the best overall efficiency. It uses 84.64M tokens, compared with 112.46M for HippoRAG2+IRCoT and 120.54M for ComoRAG, corresponding to reductions of 24.74\% and 29.78\%, respectively. \docevo also reduces initial query time by 57.58\% over HippoRAG2+IRCoT and 81.80\% over ComoRAG. 
These results show that \docevo improves answer quality 
while reducing both total token usage and initial query time.

\section{Conclusion}
\label{sec:conclusion}
This paper presents \docevo, a multi-agent RAG framework for long-document QA.
By introducing query-triggered Hypergraph-structured Working Memory that integrates document content on demand during reasoning, a Document Structural Tree Index that preserves the document's inherent hierarchy and a Graph-structured Experience Memory that stores successful reasoning plans, \docevo enables efficient, document-structure-aware and experience-guided reasoning over long-document QA tasks. 
Experimental results on four long-document QA datasets show that \docevo achieves the best overall performance, outperforming the strongest baseline, ComoRAG, with average relative gains of 16.91\% in F1 and 15.50\% in EM on open-form QA benchmarks, and by up to 20.67\% and 23.78\%, respectively, on NarrativeQA.



\twocolumn

\bibliography{aaai2027}


\end{document}